\documentclass[journal]{IEEEtran}

\usepackage{cite}
\usepackage{url}
\usepackage{optidef}
\usepackage{graphicx}
\usepackage{float}
\usepackage{graphicx}
\usepackage{subfig}
\usepackage{mathrsfs}
\usepackage{float}
\usepackage{amsmath}
\usepackage{amsfonts}
\usepackage{booktabs}
\usepackage{multirow}
\usepackage{multirow}

\usepackage{bm}

\DeclareMathOperator*{\argmax}{arg\,max}

\usepackage{algorithm}
\usepackage{algpseudocode}

\begin{document}

\title{Individual Mobility Prediction: An Interpretable Activity-based Hidden Markov Approach}
%
%
%

\author{Baichuan~Mo,~\IEEEmembership{}
        Zhan~Zhao*,~\IEEEmembership{}
        Haris N.~Koutsopoulos,~\IEEEmembership{}
        and Jinhua~Zhao~\IEEEmembership{}
\thanks{Baichuan Mo is with the Department
of Civil and Environmental Engineering, Massachusetts Institute of Technology, Cambridge, MA, 02139 USA}
\thanks{Zhan Zhao is with the Department of Urban Planning and Design, The University of Hong Kong, Hong Kong, China}
\thanks{Haris N. Koutsopoulos is with the Department of Civil and Environmental
Engineering, Northeastern University, Boston, MA 02115 USA}
\thanks{Jinhua Zhao is with the Department
of Urban Studies and Planning, Massachusetts Institute of Technology, Cambridge, MA, 02139 USA}
}

\markboth{Submitted to IEEE Transactions on Intelligent Transportation Systems}%
{Mo \MakeLowercase{\textit{et al.}}: Individual Mobility Prediction: An Activity-based Hidden Markov Method with Interpretability}

\maketitle

\begin{abstract}
Individual mobility is driven by demand for activities with diverse spatiotemporal patterns, but existing methods for mobility prediction often overlook the underlying activity patterns. To address this issue, this study develops an activity-based modeling framework for individual mobility prediction. Specifically, an input-output hidden Markov model (IOHMM) framework is proposed to simultaneously predict the (continuous) time and (discrete) location of an individual's next trip using transit smart card data. The prediction task can be transformed into predicting the hidden activity duration and end location. Based on a case study of Hong Kong's metro system, we show that the proposed model can achieve similar prediction performance as the state-of-the-art long short-term memory (LSTM) model. Unlike LSTM, the proposed IOHMM model can also be used to analyze hidden activity patterns, which provides meaningful behavioral interpretation for why an individual makes a certain trip. Therefore, the activity-based prediction framework offers a way to preserve the predictive power of advanced machine learning methods while enhancing our ability to generate insightful behavioral explanations, which is useful for enhancing situational awareness in user-centric transportation applications such as personalized traveler information.
\end{abstract}


\IEEEpeerreviewmaketitle

\section{Introduction}\label{intro}

\IEEEPARstart{I}{ndividual} mobility prediction describes the prediction of human movements over space and time at the individual level. It has important smart city and smart transportation applications, including personalized traveler information, targeted demand management, etc. Despite the emergence of extensive urban data, it is a challenging problem to accurately predict individual mobility. Travel behavior concerns multiple dimensions (most notably the temporal and spatial dimensions), exhibits longitudinal variability for an individual, and varies across individuals \cite{goulet-langlois_measuring_2017}, making the mobility prediction problem difficult to tackle. 

Individual mobility prediction is complex and multi-dimensional. While the literature mostly focuses on the problem of next location prediction \cite{calabrese_human_2010,lu_approaching_2013,hawelka_collective_2017,alhasoun_city_2015}, relatively less attention was given to the problem of next trip prediction. In a prior related study, \cite{zhao_individual_2018} defined several sub-problems related to the next trip prediction problem. It is found that, while it is easier to predict whether an individual travels or not, it is much harder to predict when and where they go next. This is not surprising because of the large number of possible combinations of people's spatiotemporal choices. It is generally challenging to deal with high-dimensional problems, especially when the data is relatively sparse (at the individual level). Besides, the existing methods are limited in that the time of travel is often treated as a categorical variable. The arbitrary discretization of time does not represent people's temporal choices adequately, and may exacerbate the data sparsity issue. Furthermore, while spatial and temporal choices of travel are typically made simultaneously, existing methods often simplify the problem to a sequential prediction task \cite{zhao_individual_2018}. This study aims to address these challenges.

The main objective of the paper is to develop a methodology to simultaneously predict the time and location of an individual's next trip. Instead of directly predicting travel behavior, we propose an input-output hidden Markov model (IOHMM) approach to analyze the underlying activity behavior. Individual mobility is driven by demand for activities with diverse spatiotemporal patterns, and thus uncovering the latter can help us predict the former. For example, the prediction of the activity duration is equivalent to that of the start time of the next trip. 

The contributions of this study are twofold. First, it enables the extraction of latent activity patterns and provides a more natural behavioral representation. This is in contrast with the approach proposed in \cite{zhao_individual_2018}, which lacks such natural behavioral interpretation. An activity-based modeling framework captures the underlying generative mechanism of travel behavior, and uncovers people's travel purposes not directly observable in the data. Second, it allows for simultaneous prediction of the discrete location and continuous time of travel. It has been shown that the temporal aspect of individual mobility is least predictable \cite{zhao_individual_2018}, and this may be partly caused by the arbitrary discretization of time. A continuous representation of time is likely to mitigate this issue. Because of the use of latent activity patterns in our proposed methodology, it allows the trip start time to be represented as a continuous emission variable. While the proposed model is general, transit smart card data from Hong Kong's Mass Transit Railway (MTR) system are used for validation.

\section{Literature review}\label{liter}

The literature on individual mobility prediction mostly focuses on the problem of next location prediction, rather than next trip prediction. Most existing methods for next location prediction are based on mining sequential patterns of individual location histories. Simple Markov chain (MC) models have shown to be able to achieve good prediction performance \cite{gambs_next_2012, lu_approaching_2013}. \cite{asahara_pedestrian-movement_2011} proposed a mixed Markov chain model (MMM) for next location prediction by identifying the group a particular individual belongs to and applying a specific MC model for that group. \cite{mathew_predicting_2012} presented a hybrid method of clustering location histories according to their characteristics before training a hidden Markov model (HMM) for each cluster. Recently, variants on Recurrent Neural Network (RNN) models have also been used for next location prediction, and showed improved prediction performance over MC models \cite{liu_predicting_2016,al-molegi_stf-rnn_2016,feng_deepmove_2018}. However, none of these methods explicitly consider the temporal behavior of the individual in the model. This is important for any mobility service because travel demand is dynamic and time-sensitive.

For the next trip prediction, we have to model the temporal behavior of individuals as well. \cite{gidofalvi_when_2012} developed a continuous-time Markov model to predict when an individual will leave their current location and where they go next. \cite{hsieh_t-gram:_2015} introduced a time-aware language model, \textit{T-gram}, to predict when an individual leaves a location by extracting location-specific time distributions from social media check-in data. More recently, \cite{zhao_individual_2018} explicitly formulated the spatiotemporal choices of individuals as a sequence of decisions, and proposed a \textit{mobility N-gram} model to predict the choices associated with the next trip---the trip start time, origin, and destination. It is found that the start time is the least predictable aspect of the next trip. While the low predictability for trip start time is to some extent rooted in people's inherent behavioral variability, the discrete representation of time is likely to limit our ability to predict temporal behavior. The key challenge is to capture the complex interaction of continuous time choices and discrete location choices. One possible approach is through latent variables representing hidden activities between trips \cite{zhao_discovering_2020}. In this work, we will extend the framework of \cite{zhao_individual_2018} and develop a methodology to capture the continuous nature of temporal behavior through latent activity patterns.

The methodology proposed in this study is based on the Input-Output Hidden Markov Model (IOHMM), which is an extension of standard HMMs \cite{bengio_input_1995}. The standard HMMs assume homogeneous transition and emission probabilities, in which the contextual information cannot be captured. To overcome this limitation, the IOHMM was proposed to incorporate additional information. Specifically, transition probabilities in IOHMMs are conditional on the input and thus depend on time. IOHMM was designed for sequence data processing, and has been applied for diverse problems including grammar inference \cite{bengio_input-output_1996}, gesture recognition \cite{marcel_hand_2000}, audio processing \cite{li_learning_2006}, and electricity price forecasting \cite{gonzalez_modeling_2005}. Our study introduces IOHMMs for individual mobility prediction. As we will demonstrate in Section~\ref{iohmm}, the IOHMM architecture can be adapted to (1) capture the dynamics of individual travel-activity histories, (2) incorporate rich contextual information for improved prediction performance, and (3) allow the modeling of both discrete (location) and continuous (time) attributes of trips/activities simultaneously.

\section{Methodology}
\subsection{Problem description}

\begin{figure}[htb]
\centering
\includegraphics[width=1 \linewidth]{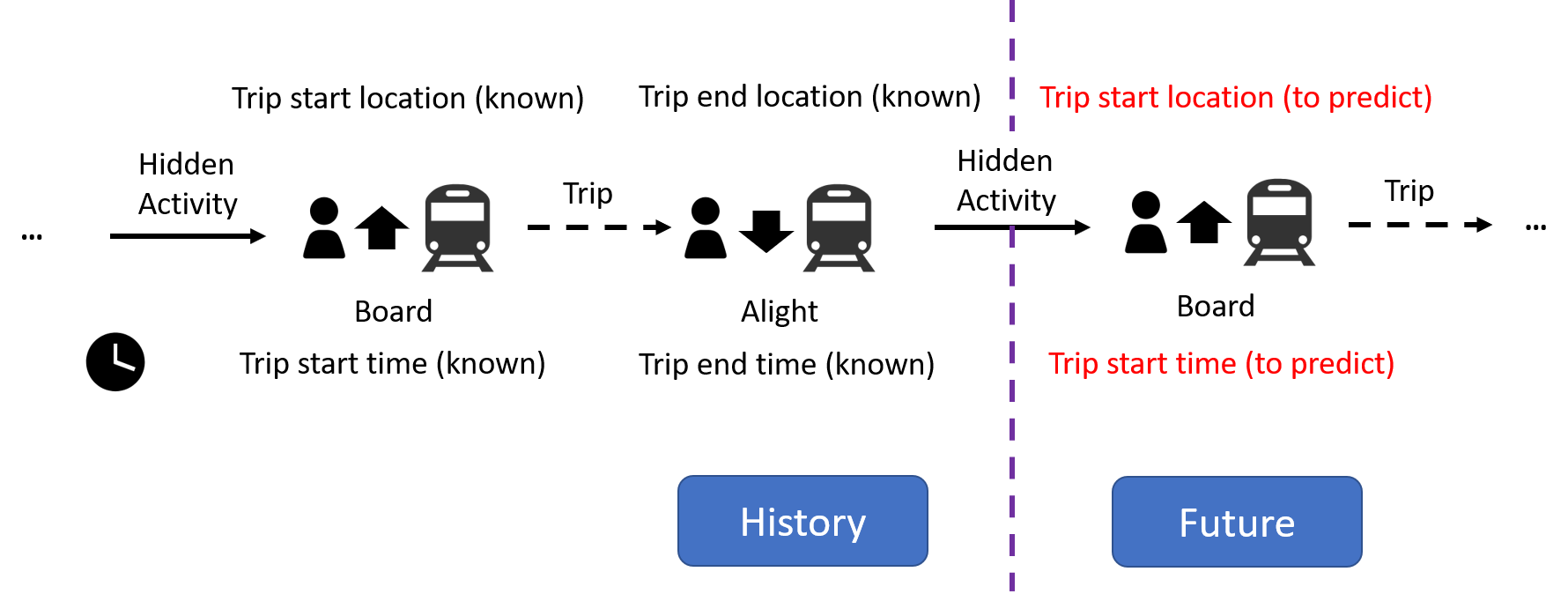}
\caption{Public transit trip structure}
\label{fig_problem}
\end{figure}

Transit smart card data, including passengers' tap-in and tap-out\footnote{This study focuses on closed public transit systems with both tap-in and tap-out records} transaction records, can provide the chronological public transit (PT) trip histories of each individual. The trip structure is shown in Figure \ref{fig_problem}. Each trip starts with boarding at an origin station and ends with alighting at a destination station. The boarding (and alighting) times and locations are known from the transit smart card data. The unique ID of each smart card allows us to track the trip histories of each anonymous individual. Between two consecutive trips, a passenger may have some activities such as working, staying at home, etc. In this study, the latent behavior of an individual between two adjacent trips is referred to as a \textit{hidden activity}. Different from a typical definition of an activity where passengers stay in a place, the hidden activity in this study may include unobserved trips such as taking a taxi to another place. Due to data limitations, we cannot identify people's trips outside the PT system. Thus, we assume that no matter what people have done between two adjacent transit trips, this process is treated as a single hidden activity. The alighting station of the last trip and the boarding station of the next trip is referred to as activity start and end locations, respectively. Our goal is to predict when and where the next trip will start given a sequence of recorded trip histories. 

Since the alighting time of the last trip is known, predicting the next trip start time is equivalent to estimating the duration of the current hidden activity. Similarly, predicting the next trip start location is equivalent to predicting the end location of the current hidden activity. In this way, we can transform the next trip prediction problem into an activity duration and location prediction problem. The new perspective has more relevant behavioral implications, as activities are usually what drive people to travel.

\subsection{Activity-based modeling framework}\label{sec_act_based}
The duration of people's hidden activities can vary greatly, anywhere from one hour (e.g. shopping) to several days (e.g. vacation). The wide range of activity duration makes it challenging to predict. In this study, we set the basic prediction interval as one day. A sequence of consecutive activities is extracted from the smart card data on a specific day and each individual may have multiple sequences of activities. The choice of prediction interval of one day not only reduces the scope of the prediction problem (from infinity to 24 hours) but also represents the basic period of regularity for human mobility and activity patterns \cite{eagle_reality_2006, kim_periodic_2007, zhao_individual_2018}. Specifically, each day spans from 4:00 AM to 4:00 AM of the next calendar day, which better matches people's daily activity schedules and the operating time of transit. 

\begin{figure}[htb]
\centering
\includegraphics[width=1 \linewidth]{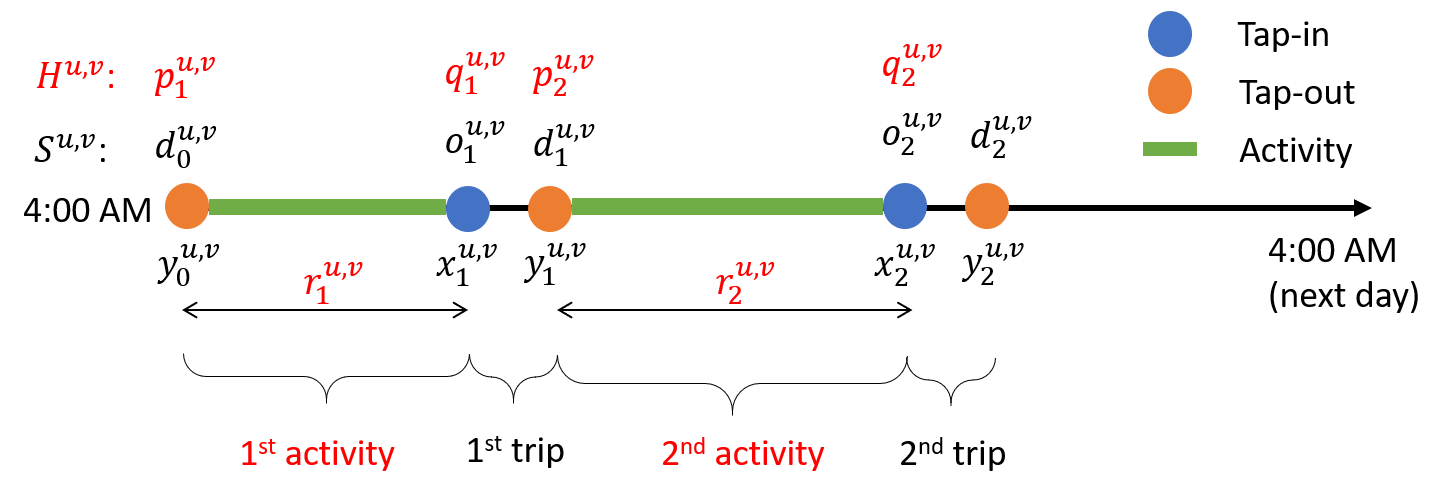}
\caption{Relationship between $S^{u,v}$ and $H^{u,v}$. There are two trips (thus two activities) in the day for this example.}
\label{fig_relation_act_trip}
\end{figure}

For a user $u$, the recorded public transit trips in day $v$ are represented as 
\begin{align}
S^{u,v} = \{(o_1^{u,v},d_1^{u,v},x_1^{u,v},y_1^{u,v}),...,(o_{T^{u,v}}^{u,v},d_{T^{u,v}}^{u,v},x_{T^{u,v}}^{u,v},y_{T^{u,v}}^{u,v}) \}
\end{align}
where $o_t^{u,v}, d_t^{u,v},x_t^{u,v},y_t^{u,v}$ are the origin, destination, start time, and end time of $t$-th trip for user $u$ in day $v$, respectively. $T^{u,v}$ is the total number of trips. The corresponding hidden activity sequence is defined as 
\begin{align}
H^{u,v} = \{(p_1^{u,v}, q_1^{u,v},r_1^{u,v}),...,(p_{T^{u,v}}^{u,v},q_{T^{u,v}}^{u,v},r_{T^{u,v}}^{u,v}) \}
\end{align}
where $p_t^{u,v}, q_t^{u,v},r_t^{u,v}$ are the start location, end location, and duration of $t$-th activity for user $u$ in day $v$, respectively. Particularly, for $t = 1,...,T^{u,v}$, we have
\begin{align}
p_t^{u,v} &= d_{t-1}^{u,v} \\ 
q_t^{u,v} &= o_{t}^{u,v} \\
r_t^{u,v} &= x_{t}^{u,v} - y_{t-1}^{u,v}
\end{align}
For the first activity, we explicitly define $d_{0}^{u,v} = \text{``null''}$ and $y_{0}^{u,v} = \text{4:00 AM}$. An example to illustrate the relationship between $S^{u,v}$ and $H^{u,v}$ is shown in Figure \ref{fig_relation_act_trip}. After a trip ends, $p_t^{u,v}$ is directly observed from the transit smart card records. Therefore, our goal is to predict $q_t^{u,v}$ and $r_t^{u,v}$ given historical trajectories and other information (e.g., weather). 

It is worth noting that we do not consider the time period from $y_{T^{u,v}}^{u,v}$ to 4:00 AM next day as the last activity interval because its duration is deterministic ($y_{T^{u,v}}^{u,v}$ is known). This study focuses on predicting the next trip's time and location, but there is no corresponding next trip for this activity. Therefore, there is no need to predict the last activity, and it is excluded from further analysis. 

\subsection{IOHMM for activity prediction}\label{iohmm}
\begin{figure}[htb]
\centering
\includegraphics[width=1 \linewidth]{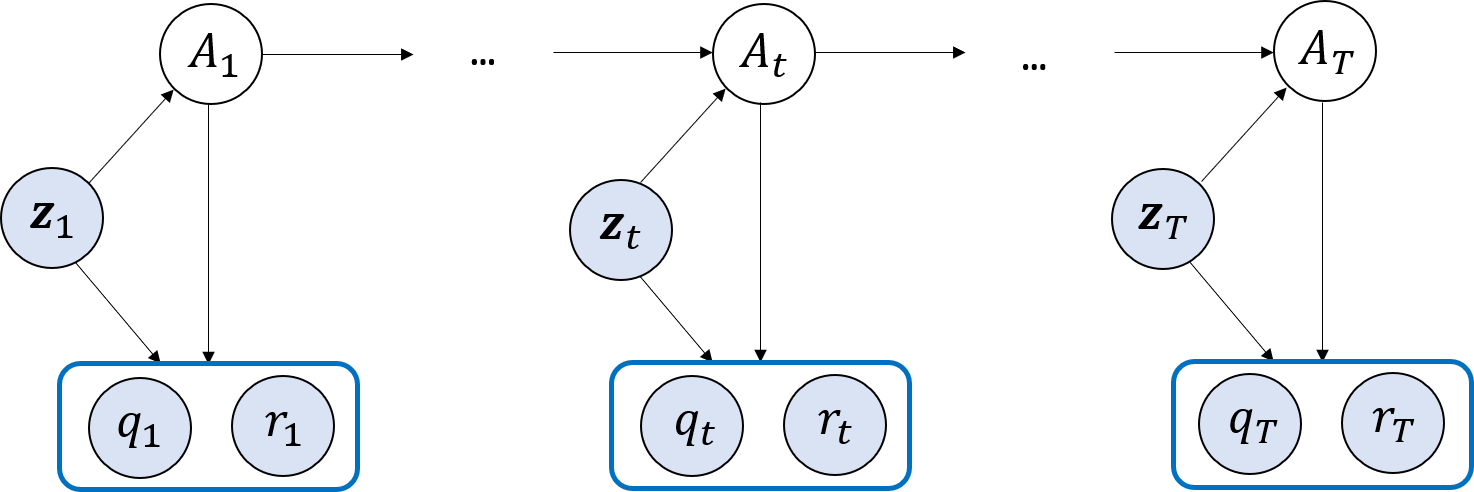}
\caption{Structure of IOHMM. The solid nodes represent observed information, while the transparent (white) nodes represent latent random variables. }
\label{fig_iohmm}
\end{figure}

IOHMM is proposed to capture exogenous contextual information over time, which allows modeling of heterogeneous transition and emission probabilities. The structure of IOHMM for individual activity modeling is shown in Figure \ref{fig_iohmm}. $A_t$ is the $t$-th hidden activity (a latent random variable) and $\bm{z_t}$ is a vector of observed input variables containing contextual information (e.g., weather, day of week, $p_t$, etc.). The superscript $(u,v)$ is ignored for simplicity. Since each hidden activity can be encoded as a latent state in IOHMM, the IOHMM architecture matches well with the activity-based modeling framework. 

The model consists of three key components: 1) initial state probability $\pi_i = \mathbb{P}(A_1 = i \;|\; \bm{z_1};\;\bm{\theta_{in}})$, where $i \in \mathcal{A}$ and  $\mathcal{A}$ is the state space; It quantifies the distribution of the first activity's type. 2) transition probability: $\varphi_{ij,t} = \mathbb{P}(A_t = j \;|\;A_{t-1} = i, \bm{z_t};\;\bm{\theta_{tr}})$, which quantifies the probability that next activity is $j$ given this activity is $i$, and 3) emission probability: $\delta_{i,t} = \mathbb{P}(q_t, r_t \;|\; A_t = i, \bm{z_t};\;\bm{\theta_{em}})$, which quantifies the distributions of activity duration and end location. $\bm{\theta_{in}}$, $\bm{\theta_{tr}}$, and $\bm{\theta_{em}}$ are parameters of initial, transition, and emission probability functions, respectively. The likelihood of a data sequence under this model is given by:
\begin{align}
L(\bm{\theta}) = &\sum_{A_1,...,A_T} \mathbb{P}(A_1\;|\; \bm{z_1};\;\bm{\theta_{in}})\cdot \prod_{t=2}^T\mathbb{P}(A_t \;|\; A_{t-1}, \bm{z_t};\;\bm{\theta_{tr}}) \cdot \nonumber\\ 
&\prod_{t=1}^T \mathbb{P}(q_t, r_t \;|\; A_t, \bm{z_t};\;\bm{\theta_{em}})
\end{align}
where $\bm{\theta} = [\bm{\theta_{in}},\bm{\theta_{tr}},\bm{\theta_{em}}]$. 

The model is estimated by the Expectation-Maximization (EM) algorithm. 

\textbf{E-step}: Denote the estimated parameters at iteration $k-1$ of M-step as $\bm{\theta}^{(k-1)}$ (if $k=1$, use the initial values of the parameters). From $\bm{\theta}^{(k-1)}$ we can obtain the three probabilities as $\pi_i^{(k-1)},\delta_{i,t}^{(k-1)},\varphi_{ij,t} ^{(k-1)}$. Then, the forward and backward variables (denoted as $\alpha_{i,t}^{(k)}$ and $\beta_{i,t}^{(k)}$, respectively) are calculated as 
\begin{align}
\alpha_{i,t}^{(k)} &= \mathbb{P}(q_{1:t},r_{1:t},A_t = i \;|\;\bm{z_{1:t}}) = \delta_{i,t}^{(k-1)} \sum_{l\in\mathcal{A}}\varphi_{li,t}^{(k-1)}\cdot\alpha_{l,t-1}^{(k)} \label{eq_forward}\\
\beta_{i,t}^{(k)} &= \mathbb{P}(q_{t+1:T},r_{t+1:T}\;|\;A_t = i, \bm{z_{t:T}}) \nonumber \\
&=  \sum_{l\in\mathcal{A}}\varphi_{il,t}^{(k-1)} \cdot \beta_{l,t+1}^{(k)} \cdot \delta_{l,t+1}^{(k-1)} 
\end{align}
where $\alpha_{i,1}^{(k)} = \pi_i^{(k-1)}  \delta_{i,1}^{(k-1)}$ and $\beta_{i,T} = 1$. The subscripts $1:t$ indicates a list of the corresponding variable with subscript from $1$ to $t$. Then, we calculate the posterior state probability and posterior transition probability as:
\begin{align}
\gamma_{i,t}^{(k)} &= \mathbb{P}(A_t = i \;|\;q_{1:T},r_{1:T},\bm{z}_{1:T}) = \alpha_{i,t}^{(k)} \cdot\beta_{i,t}^{(k)} / L_c^{(k)} \\
\xi_{ij,t}^{(k)} &= \mathbb{P}(A_t = j, A_{t-1} = i\;|\;q_{1:T}, r_{1:T}, \bm{z}_{t:T}) \nonumber\\
&= \varphi_{ij,t}^{(k-1)} \cdot \alpha_{i,t-1}^{(k)} \cdot \beta_{j,t}^{(k)} \cdot \delta_{j,t}^{(k-1)} / L_c^{(k)}
\end{align}
where $L_c^{(k)}$ is the complete data likelihood at iteration $k$, defined as $L_c = \sum_{i\in\mathcal{A}} \alpha_{i,T}^{(k)}$. Obtaining $\alpha_{i,t}^{(k)}$, $\beta_{i,t}^{(k)}$, $\gamma_{i,t}^{(k)}$, and $\xi_{ij,t}^{(k)}$ for all $i,j\in\mathcal{A}$ and $t=1,...,T$ finishes the E-step. 

\textbf{M-step}: The probability parameters in iteration $k$ is updated as by maximizing the expected data log likelihood: 
\begin{align}
Q(\bm{\theta} ;\; \bm{\theta}^{(k-1)}) &= \sum_{i\in \mathcal{A}} \gamma_{i,1}^{(k)}\cdot\log \mathbb{P}(A_1 = i\;|\;\bm{z_1}; \;\bm{\theta_{in}}) \nonumber \\
&+\sum_{t=2}^T\sum_{i,j\in\mathcal{A}}\xi_{ij,t}^{(k)}\cdot\log \mathbb{P}(A_t = j\;|\;A_{t-1} = i, \bm{z_t}; \;\bm{\theta_{tr}}) \nonumber\\
&+\sum_{t=1}^T\sum_{i\in\mathcal{A}}\gamma_{i,t}^{(k)}\cdot\log \mathbb{P}(q_t,r_t \;|\;A_{t-1} = i, \bm{z_t}; \;\bm{\theta_{em}}) 
\end{align}
We have $\bm{\theta}^{(k)} = \argmax_{\bm{\theta}} Q(\bm{\theta} ;\; \bm{\theta}^{(k-1)})$. The M-step can be implemented by any supervised learning model that supports gradient ascent on the log probability. With proper specification of three probability functions, the optimization problem can be convex and easily solved. It is also worth noting that since $Q(\bm{\theta} ;\; \bm{\theta}^{(k-1)})$ consists of three components with independent parameters, maximizing  $Q(\bm{\theta} ;\; \bm{\theta}^{(k-1)})$ is equivalent to maximizing the three components separately, which allows for more flexibility in optimization. 

\subsection{Model specification}\label{model_speciciation}
\textbf{Contextual information}: In terms of the contextual information $\bm{z_t}$, five different dimensions are considered: weather, day of the week, holidays, last trip information, and historical travel statistics. The specific variables used can be found in the case study. 

\textbf{State space}: The state space $\mathcal{A}$ is specified for each individual (i.e., $\mathcal{A}^{u}$ for user $u$). Since an activity label is a latent categorical variable during the modeling process, we only need to define the cardinality $N^{u}$, which follows that $\mathcal{A}^{u} = \{1,...,N^{u}\}$. $N^{u}$ indicates how many hidden activities we considered for user $u$. A semantic label can be associated to each element in $\mathcal{A}^{u}$ with an in-depth analysis in Section \ref{case_study}. Generally, the value of $N^{u}$ can be determined using the validation data set \cite{chiappa2003hmm} or optimization approaches \cite{geiger2010optimizing}. In this study, we select $N^{u}$ by maximizing the \emph{silhouette coefficient} \cite{rousseeuw1987silhouettes}. We assume hidden activities can be characterized by $\bm{z_t}^{u,v}$. For user $u$, we can cluster $\bm{z_t}^{u,v}$ into $m$ clusters, representing $m$ possible hidden activities. The silhouette coefficient of the $m$-clustering (denoted as $SC^{u}(m)$ for user $u$) is a measure of how similar each object is to its own cluster compared to other clusters (i.e., the quality of clustering). It is defined as
\begin{align}
SC^{u}(m) = \text{mean}\{\frac{b(i) - a(i)}{\max\{a(i),b(i)\}}\}
\end{align}
where $a(i)$ and $b(i)$ are the intra-cluster distance and nearest-cluster (that $i$ does not belong to) distance of data point $i$, respectively. ``$\text{mean}\{\}$'' indicates taking the average over all samples. $SC^{u}(m)$ ranges from -1 to +1, where a high value indicates that the samples are well matched to their own cluster and poorly matched to neighboring clusters. Hence, $N^{u}$ is obtained by
\begin{align}
N^{u} = \argmax_{m\in\mathcal{M}} SC^{u}(m)
\end{align}
where $\mathcal{M}$ is the set of possible numbers of hidden activities. In this study, $\mathcal{M} = \{3,4,...,7\}$ is used. It worth noting that we also tested other cluster quality metrics, such as Akaike information criterion (AIC) and Bayesian information criterion (BIC). Numerical results show that the silhouette coefficient works best for determining the number of hidden activities. 

\textbf{Three probability functions}: The multinomial logistic regression is used to model the initial probability and transition probability. Specifically, we have:
\begin{align}
\mathbb{P}(A_1 = i \;|\; \bm{z_1};\;\bm{\theta_{in}}) = \frac{\exp({\bm{\theta_{in,i}} \cdot \bm{z_1}})}{\sum_{j\in{\mathcal{A}}} \exp({\bm{\theta_{in,j}} \cdot \bm{z_1}})}
\end{align}
where $\bm{\theta_{in,i}}$ are the coefficients of the initial state probability function at state $i$. 
\begin{align}
\mathbb{P}(A_t = j \;|\;A_{t-1}=i, \bm{z_t};\;\bm{\theta_{tr}}) = \frac{\exp({\bm{\theta_{tr,ij}} \cdot \bm{z_t}})}{\sum_{j\in{\mathcal{A}}} \exp({\bm{\theta_{tr,ij}} \cdot \bm{z_t}})}
\end{align}
where $\bm{\theta_{tr,ij}}$ are the coefficients of the state transition probability function when the next state is $j$ given the current state is $i$.

In terms of the emission probability, it is worth noting that $q_t$ is a discrete random variable while $r_t$ continuous. We assume a conditional independence between $q_t$ and $r_t$, that is, 
\begin{align}
&\mathbb{P}(q_t, r_t \;|\; A_t = i, \bm{z_t};\;\bm{\theta_{em}})  \nonumber \\
&=\mathbb{P}(q_t \;|\; A_t = i, \bm{z_t};\;\bm{\theta_{emq}}) \cdot \mathbb{P}(r_t \;|\; A_t = i, \bm{z_t};\;\bm{\theta_{emr}})
\end{align}
For the activity end location distribution, a similar multinomial logistic regression model is used, where
\begin{align}
\mathbb{P}(q_t = l \;|\; A_t = i, \bm{z_t};\;\bm{\theta_{emq}}) = \frac{\exp({\bm{\theta_{emq,i,l}} \cdot \bm{z_t}})}{\sum_{l\in{\mathcal{L}}} \exp({\bm{\theta_{emq,i,l}} \cdot \bm{z_t}})}
\end{align}
where $\bm{\theta_{emq,i,l}}$ are the coefficients for emission probability of activity location where the location is $l$ given the current state is $i$. $\mathcal{L}$ is the set of location candidates. For user $u$, $\mathcal{L}^{u}$ is defined as all stations that he/she has visited in the smart card data records. In terms of the duration distribution, we assume a
Gaussian distribution with the mean quantified by a linear function:
\begin{align}
\mathbb{P}(r_t \;|\; A_t = i, \bm{z_t};\;\bm{\theta_{emr}}) = \frac{1}{\sqrt{2\pi}\sigma_i}e^{-\frac{(r_t - \bm{\theta_{emr,i}}\cdot  \bm{z_t})^2 }{2\sigma_i^2}}
\end{align}
where $\bm{\theta_{emr,i}}$ and ${\sigma_i}$ denote the coefficients and the standard deviation of the model when the hidden state is $i$.

\subsection{Prediction formulation}\label{prediction}
The IOHMM supports predicting the next activity duration and location given today's trajectories after training. Here we only give the formulation for predicting the duration, and that for location prediction can be derived in the same way. Observe that predicting the duration of the next activity is equivalent to obtaining $\mathbb{P}(r_{t+1} \;|\; q_{1:t}, r_{1:t}, \bm{z_{1:t+1}})$. This is because $q_{1:t}, r_{1:t}, \bm{z_{1:t+1}}$ are all observed information. By the conditional independence, we have $\mathbb{P}(r_{t+1} \;|\; q_{1:t}, r_{1:t}, \bm{z_{1:t+1}}) = \mathbb{P}(r_{t+1} \;|\; r_{1:t}, \bm{z_{1:t+1}})$. By the law of total probability:
\begin{align}
&\mathbb{P}(r_{t+1} \;|\; r_{1:t}, \bm{z_{1:t+1}}) \nonumber \\
&= \sum_{i \in \mathcal{A}}\mathbb{P}(r_{t+1} \;|\; A_{t+1} = i, \bm{z_{t}}) \cdot \mathbb{P}(A_{t+1} = i\;|\;r_{1:t}, \bm{z_{1:t+1}})
\label{eq_pred_1}
\end{align}
The first term in the right hand side (RHS) of Eq. \ref{eq_pred_1} is the emission probability. And the second term can be expanded as:
\begin{align}
\mathbb{P}(A_{t+1} = i\;|\;r_{1:t}, \bm{z_{1:t+1}}) = \frac{\mathbb{P}(A_{t+1} = i, r_{1:t} \;|\; \bm{z_{1:{t+1}}})}{\sum_{j \in \mathcal{A}}\mathbb{P}(A_{t+1} = j,  r_{1:t} \;|\;\bm{z_{1:t+1}})} 
\label{eq_pred_2}
\end{align}
where
\begin{align}
&\mathbb{P}(A_{t+1} = i, r_{1:t} \;|\; \bm{z_{1:{t+1}}})  \nonumber \\
&= \sum_{i \in \mathcal{A}}\mathbb{P}(A_{t+1} = j \;|\; A_{t} = i, \bm{z_{t+1}}) \cdot \mathbb{P}(A_{t} = i, r_{1:t}\;|\;\bm{z_{1:{t}}})
\label{eq_pred_3}
\end{align}
The first term in the RHS of Eq. \ref{eq_pred_3} is the state transition probability. And the second term is essentially the forward variable (Eq. \ref{eq_forward}) when only incorporating the emission probability of duration. Therefore, based on the forward variable, state transition probability, and emission probability, one can output the distribution of $r_{t+1}$ given $ r_{1:t}, \bm{z_{1:t+1}}$, which is Gaussian distributed based on our specification. We adopt the mean as the predicted duration. 

For the location distribution, we can derive $\mathbb{P}(q_{t+1} \;|\; q_{1:t}, \bm{z_{1:t+1}}) $ using the same method. The location with the highest probability is selected as the prediction.

\subsection{Model interpretability}\label{sec_interp}
The IOHMM allows us to explore the mobility patterns of an individual. In the following discussion, the subscript $t$ is ignored as we focus on deriving the general pattern over history.

\textbf{Activity pattern identification}: To identify the latent activity, four distributions conditioning on a specific activity label are calculated: 1) duration distribution $\mathbb{P}(r \;|\; A = i)$, 2) end location distribution $\mathbb{P}(q \;|\; A = i)$, 3) start time (i.e., last trip end time) $\mathbb{P}(y \;|\; A = i)$, and 4) start location distribution $\mathbb{P}(p \;|\; A = i)$. Since the transition matrix describes how passengers moving from one activity to another, we are also curious about $\mathbb{P}(A_t = j \;|\; A_{t-1} = i)$ for all $i,j\in\mathcal{A}$. Based on these distributions, we can assign a semantic label (e.g., home, work) to each hidden activity manually.  

These distributions/parameters are calculated based on Gibbs sampling as illustrated in Algorithm \ref{alg_interpre}. It is worth noting that we assume $\mathbb{P}(\bm{z})$ is the same as the distribution of $\bm{z}$ in historical trajectories. Hence, instead of sampling $\bm{z} \sim \mathbb{P}(\bm{z})$, we can generate samples by going through all histories (i.e. $t=1, ..., T^{u,v}$, for $v=1, ..., V^u$, where $V^u$ is the total number of travel days for user $u$. This can be seen as bootstrapping). The sampling process is repeated $N$ times. And the intended distributions can be directly obtained from the generated sequences (e.g., for a discrete variable, we can directly count the conditional frequency in the generated sequences).

\begin{algorithm}
\small
\caption{Activity pattern identification for a user $u$ using Gibbs sampling}
\begin{algorithmic}[1]
\renewcommand{\algorithmicrequire}{\textbf{Input:}}
\renewcommand{\algorithmicensure}{\textbf{Output:}}
\Require Trained IOHMM; Trip history of user $u$ 
\Ensure  Intended probability distribution of user $u$ 
\State Initialize the number of sampling $N$. 
\For {$n = 1$ to $N$}
\For {$v = 1$ to $V^u$}
\State Sample $A_1^{u,v} \sim \mathbb{P}(A_1^{u,v} \;|\; \bm{z_1^{u,v}})$
\State Sample $r_1^{u,v} \sim \mathbb{P}(r_1^{u,v}\;|\; A_1^{u,v}, \bm{z_1^{u,v}})$ and $q_1^{u,v} \sim \mathbb{P}(q_1^{u,v}\;|\; A_1^{u,v}, \bm{z_1^{u,v}})$.
\For {$t = 2$ to $T^{u,v}$}
\State Sample $A_t^{u,v} \sim \mathbb{P}(A_t^{u,v} \;|\;A_{t-1}^{u,v}, \bm{z_t^{u,v}})$
\State Sample $r_t^{u,v} \sim \mathbb{P}(r_t^{u,v}\;|\; A_t^{u,v}, \bm{z_t^{u,v}})$ and $q_t^{u,v} \sim \mathbb{P}(q_t^{u,v}\;|\; A_t^{u,v}, \bm{z_t^{u,v}})$.
\EndFor
\EndFor
\State Save the generated activity sequences and corresponding contextual information in iteration $n$ as $H^{u}_n$ and $\bm{z}^{u}_n$, respectively.
\EndFor
\State Obtain the intended distribution described above for user $u$ based on $[(H^{u}_1, \bm{z}^{u}_1),...,(H^{u}_N, \bm{z}^{u}_N)]$.
\end{algorithmic} 
\label{alg_interpre}
\end{algorithm}
 
\textbf{Probability coefficients explanation}: After training the model, we can obtain $\bm{\theta}$ for each probability function. Since all functions adopt a linear relationship between $\bm{\theta}$ and $\bm{z}$, the value of $\bm{\theta}$ enables interpretability and validation of the training results. For example, we may expect rain to have a positive effect on the duration of all activities.

\section{Case study}\label{case_study}
\subsection{Data}
The dataset used for the case study contains transit smart card records from 500 anonymous users between July 2014 and March 2017 in the Hong Kong Mass Transit Railway (MTR) system. These users are selected randomly from all individuals with at least 300 active days of transit usage during the study period, which excludes occasional users and short-term visitors such as tourists. This is because a minimum amount of personal travel history is required to achieve reasonable prediction performance. The mobility prediction for infrequent users and short-term visitors requires future research. It is worth noting that though an individual may hold more than one smart card and a smart card data may represent multiple users, we assume that each card ID corresponds to only one user \cite{zhao_individual_2018}. 

We partition the personal daily activity sequences of each user into training and test sets. The test set consists of the sequences from 20\% randomly selected active days. The remaining sequences form the training set. The IOHMM model is specified for each user based on their own training data. 

\subsection{Travel patterns}
\begin{figure}[htb]
\centering
\includegraphics[width=1 \linewidth]{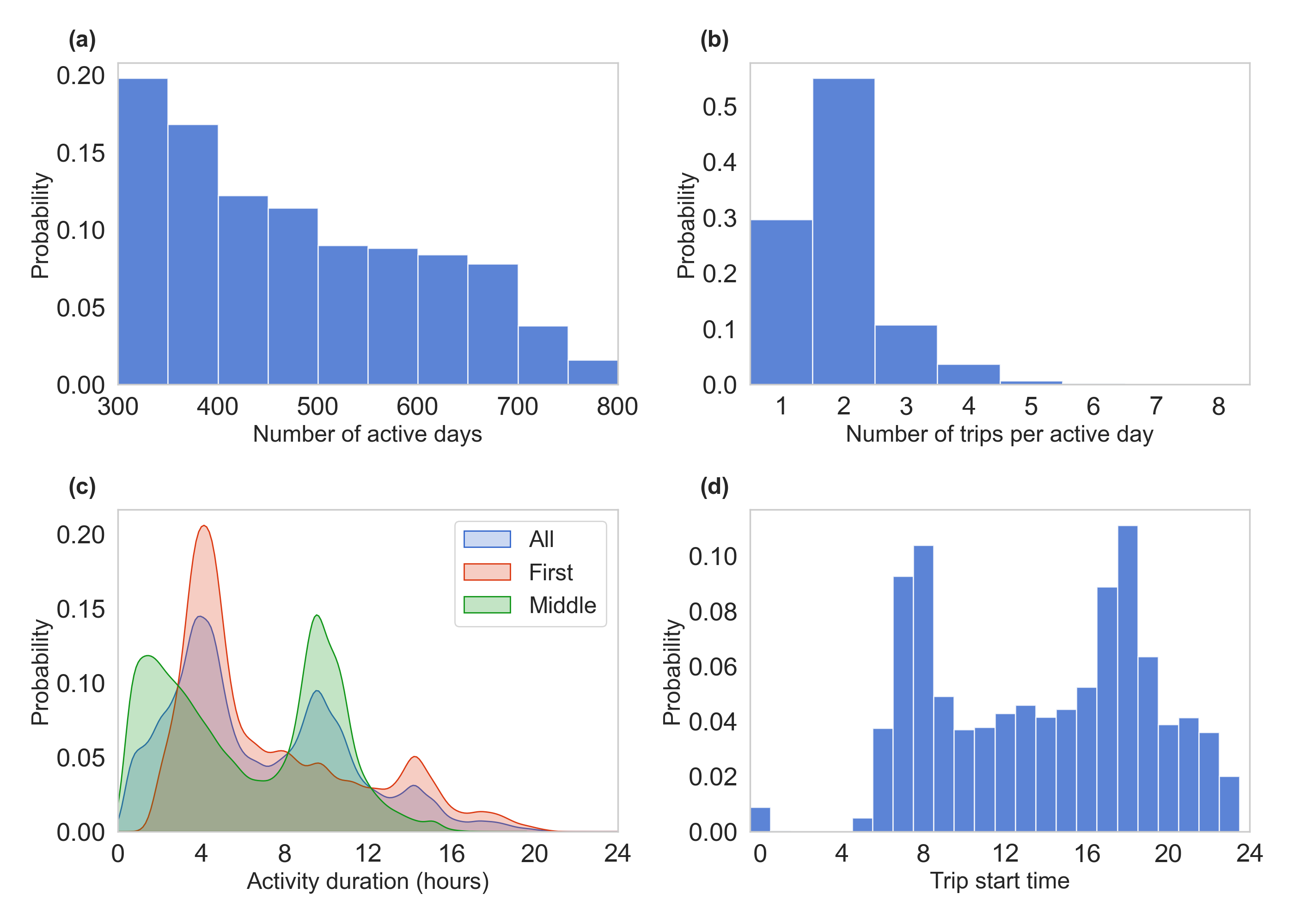}
\caption{Distribution of individual travel characteristics. In (c), ``first'' means the duration distribution of the first activity in a day. ``Middle'' means all activities in a day excluding the first one. ``All'' means all activities.}
\label{fig_sample_pattern}
\end{figure}

The travel patterns of selected sample individuals are shown in Figure \ref{fig_sample_pattern}. Figure \ref{fig_sample_pattern}(a) shows the distribution of the number of active days (i.e. days with at least one trip). We observe most of the samples have less than 400 active days during the 2.5 years of the analysis period. Figure \ref{fig_sample_pattern}(b) shows the distribution of the number of trips per active day. An individual typically makes two trips in a day, likely as a result of commuting to and from work. Note that only rail-based trips are considered in the case study. Thus, this distribution is an underrepresentation of the true travel intensity of users. The distribution of activity duration is shown in Figure \ref{fig_sample_pattern}(c). For the first activity, we observe a prominent peak of around 4 hours. This may represent the weekday ``staying at home'' activity because the start of a day is set as 4:00 AM and people usually leave home for work at around 8:00 AM. There is a sub-peak at around 14 hours for the first activity, which may correspond to the holiday ``staying at home'' activity where people stay at home until 18:00 and then leave home for leisure. For the middle activity, a major peak at around 10 hours is observed, which may indicate the work activity (start at 8:00 and end at 18:00). Another peak for middle activity is around 2 hours, which may represent short-term dining/entertainment activities. The trip start time distribution is shown in Figure \ref{fig_sample_pattern}(d), as expected, a morning peak at 8:00 AM and an evening peak at 18:00 are observed.

\subsection{Evaluation metrics and benchmark models}
Recall that the proposed IOHMM can output the predicted activity duration ($r_t^{u,v}$) and end location ($q_t^{u,v}$). $q_t^{u,v}$ is a categorical variable and the prediction accuracy is used for the performance evaluation. $r_t^{u,v}$ is a continuous variable. The predicted $R^2$ (i.e. $R^2$ in the test data set) is used as the main performance metric because it typically ranges between 0 and 1, which is consistent with the range of prediction accuracy. 

To properly evaluate the proposed IOHMM, we compare it against two types of models for benchmarking. The first group includes simple and straightforward models and can be seen as a ``lower-bound'' of the prediction performance. The second group is based on more advanced machine learning methods that are commonly used for sequential prediction. It can be seen as providing an approximate ``upper-bound'' of the prediction performance for existing approaches. Specifically, linear regression (LR) and the first-order Markov Chain (MC) model are used as the first type benchmark models for predicting $r_t^{u,v}$ and $q_t^{u,v}$, respectively. LR is used because it is the most commonly used model for continuous variable prediction. The MC model is used because it was shown in \cite{lu2013approaching} that the first-order MC can approach the limit of predictability for the next location prediction and it was previously used in \cite{zhao_individual_2018} as the baseline model for location prediction. 

The LR model for user $u$ is formulated as
\begin{align}
r_t^{u,v} = \beta_0^u + \beta^u \cdot \bm{z_t^{u,v}} + \epsilon^u  \quad\quad\; \forall v,t
\end{align}
where $\beta_0^u$ is the intercept and $\beta^u$ is the vector of parameters to estimate. $\epsilon^u$ is the error term. 

In terms of the MC model, the distribution of the activity end location (i.e. next trip origin) is formulated as 
\begin{align}
\mathbb{P}(q_1^{u,v}) &= \mathbb{P}(o_1^{u,v}) = \frac{C(o_1^{u,v}) + \alpha/|\mathcal{L}^u|}{V^u + \alpha}  \\
\mathbb{P}(q_t^{u,v}\;|\;p_t^{u,v}) &= \mathbb{P}(o_t^{u,v}\;|\;d_{t-1}^{u,v}) \nonumber \\
&= \frac{C(d_{t-1}^{u,v}, o_t^{u,v}) + \alpha/|\mathcal{L}^u|}{\sum_{\tilde{o}_t^{u,v} \in{\mathcal{L}^u}} C(d_{t-1}^{u,v}, \tilde{o}_t^{u,v}) + \alpha} \quad\quad\; \forall t \geq 2
\end{align}
where $C(o_1^{u,v})$ is a counting function that returns the number of times that the first trip of the day starts at $o_1^{u,v}$. Similarly, $C(d_{t-1}^{u,v}, o_t^{u,v})$ returns the number of times that a trip ending at $d_{t-1}^{u,v}$ is followed (in the same day) by another trip starting from $o_t^{u,v}$. $\mathcal{L}^u$ is the set of candidates location for user $u$. The parameter $\alpha$ is used for smoothing so that a non-zero probability is generated for any
possible value.

\begin{figure}[htb]
\centering
\includegraphics[width=0.7 \linewidth]{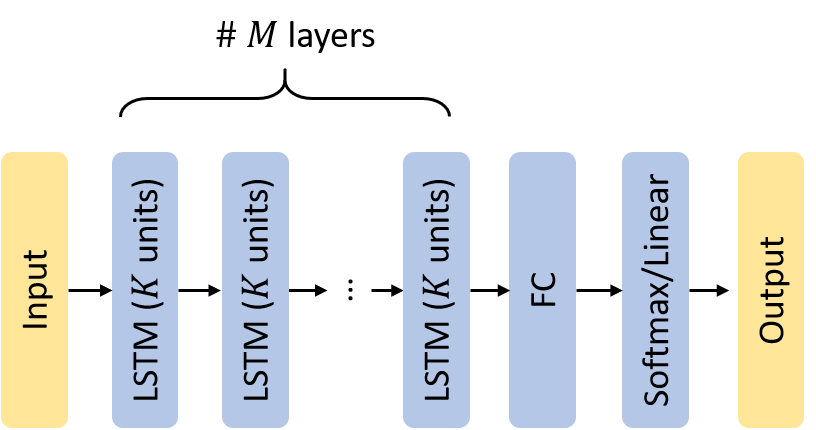}
\caption{LSTM network structure. $M$ is the number of LSTM layers. $K$ is the number of hidden units. ``FC'' means fully connected neural network layer}
\label{fig_lstm}
\end{figure}

Long short-term memory (LSTM) \cite{hochreiter1997long} is selected to represent the second type of benchmark models. LSTM is a recurrent neural network (RNN) architecture used in the field of deep learning.  It is well-suited to classify, process, and predict time series given time lags of unknown duration as it has the advantage of memorizing long-range dependencies in the data. LSTM has been widely used in different tasks such as handwriting recognition \cite{graves2008novel} and speech recognition \cite{sak2014long}, and is often considered as one of the state-of-the-art methods for time series prediction tasks. Since LSTM is not suitable for predicting continuous ($r_t^{u,v}$) and discrete ($q_t^{u,v}$) variables simultaneously, we train  for each individual two separate LSTM models to predict $r_t^{u,v}$ and $q_t^{u,v}$, respectively. The structure of LSTM is shown in Figure \ref{fig_lstm}. Input variables are fed into $M$ LSTM layers and then aggregated by a fully connected (FC) neural network layer. For duration (resp. location) prediction, the linear (resp. softmax) activation layer is used for the outputs. The hyper-parameters (e.g., number of LSTM layers $M$, number of hidden units $K$, regularization strength, drop out rate) are tuned based on a searching process over a predetermined hyper-parameter space (see Appendix \ref{sec_hyper}). A validation data set (20\% of the training data) is used for the hyper-parameters selection. The hyper-parameters with the highest $R^2$ and prediction accuracy in the validation data are used as the final models. 

\subsection{Prediction performance}
\begin{figure}[htb]
\centering
\subfloat[$R^2$ distribution for activity duration prediction]{\includegraphics[width=1\linewidth]{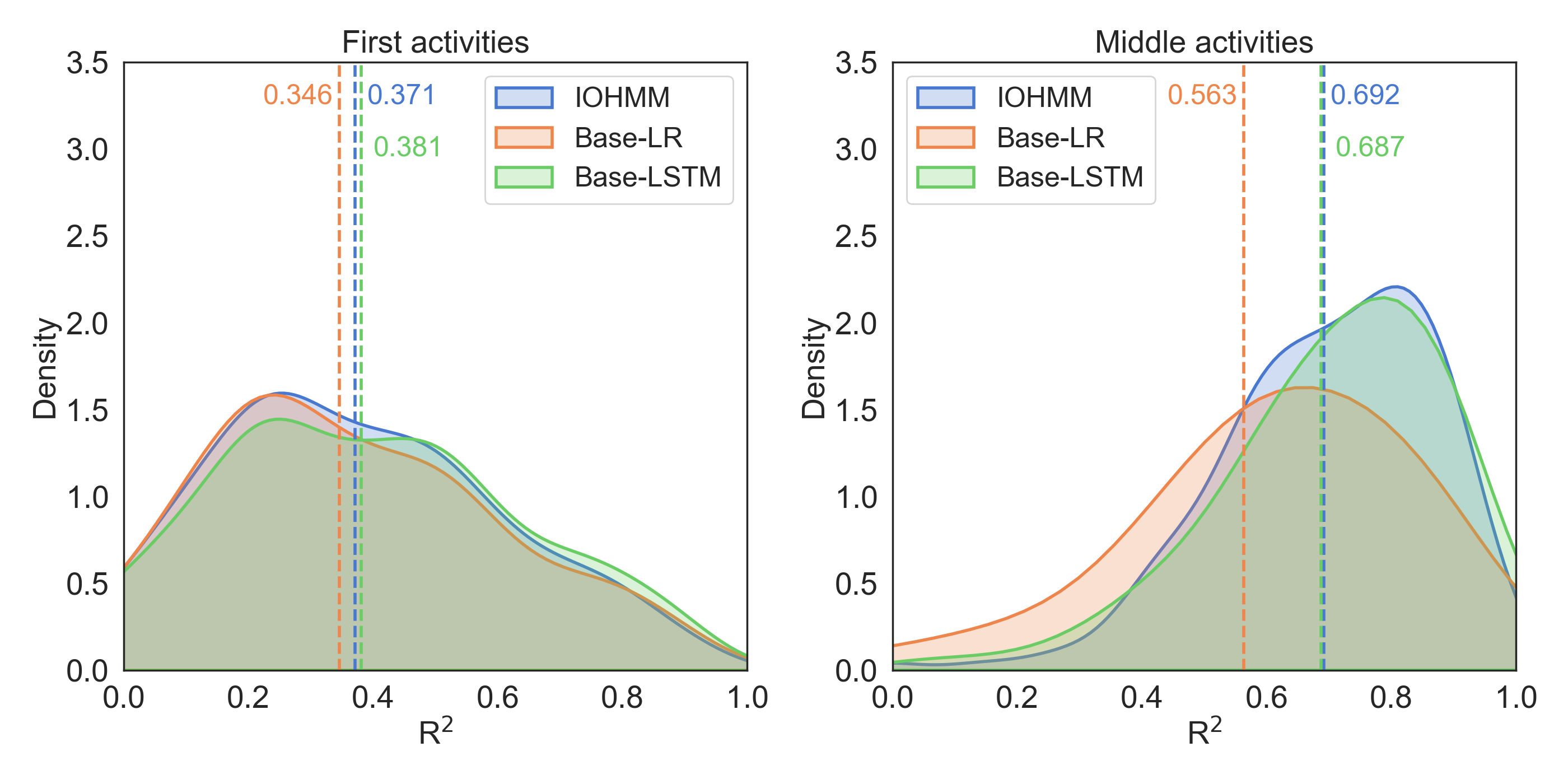}\label{fig_station_flow_1}}
\hfil
\subfloat[Accuracy distribution for activity end location prediction]{\includegraphics[width=1\linewidth]{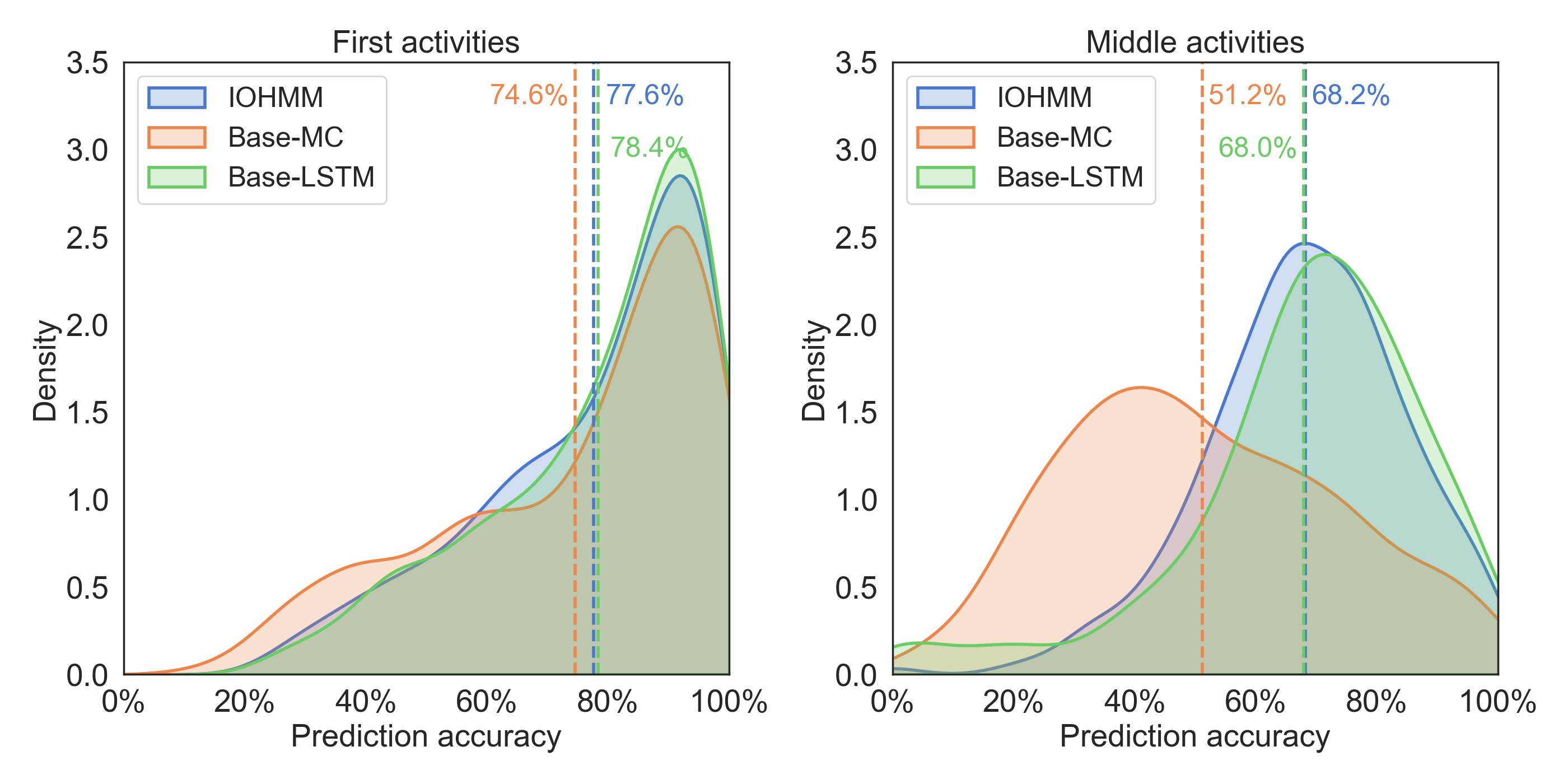}\label{fig_station_flow_2}}
\caption{Prediction performance. ``First activities'' means the first activity in a day. ``Middle activities'' means all activities in a day excluding the first one. The dash lines represent the mean value.}
\label{fig_pred}
\end{figure}

As each model outputs $R^2$ and prediction accuracy for each individual, we can plot the distribution of $R^2$ and prediction accuracy for overall performance evaluation. Figure \ref{fig_pred} shows the prediction performance for the activity duration and location. Since the first activities are predicted by the initial probability and the middle activities are predicted by the transition probability, we plot the performance distribution for two types of activities separately. What stands out in the figure is a high degree of individual heterogeneity in terms of predictability. Overall, the IOHMM shows very similar performance as the LSTM model in all prediction tasks. And both IOHMM and LSTM can outperform the first type of baseline models (i.e. LR and MC). This implies that the proposed IOHMM not only has the same predictive capacity as the advanced machine learning model but also has the potential to identify latent activities with model interpretability (details illustrated in Section \ref{sec_latent_act}). 

In terms of the duration prediction (Figure \ref{fig_pred}a), we observe that IOHMM and LSTM are only slightly better than LR (with mean $R^2=0.371, 0.381$, and $0.346$, respectively). This implies that people's first trip start time on a day has high randomness and is hard to predict as many uncaptured reasons can cause morning departure times to be adjusted. However, for the middle activities, the IOHMM and LSTM significantly outperform the LR model (with mean $R^2=0.692, 0.687$, and $0.563$, respectively).

The results for location prediction (Figure \ref{fig_pred}b) are similar to those of duration prediction. IOHMM and LSTM models are slightly better than the MC model in the first activity end location prediction, but significantly better in the prediction of middle activities. An interesting finding is that, though the duration of the first activity is relatively difficult to predict, the prediction accuracy for the first activity end location (i.e. first trip origin) is high (with a mean of 77.6\%, 78.4\%, and 74.6\% for IOHMM, LSTM, and MC, respectively). This implies that despite randomness in start time, the first trip origins are relatively stable for these frequent public transit users. The location prediction accuracy for middle activities is lower than that of the first activities (with a mean of 68.2\%, 68.0\%, and 51.2\% for IOHMM, LSTM, and MC, respectively). This may be attributed to the higher degree of behavioral randomness after leaving home. For the first activity, people are likely to use the nearest rail station around the home. But for middle activities, people may have more choices which are not easy to capture. 

\begin{figure}[htb]
\centering
\includegraphics[width=1 \linewidth]{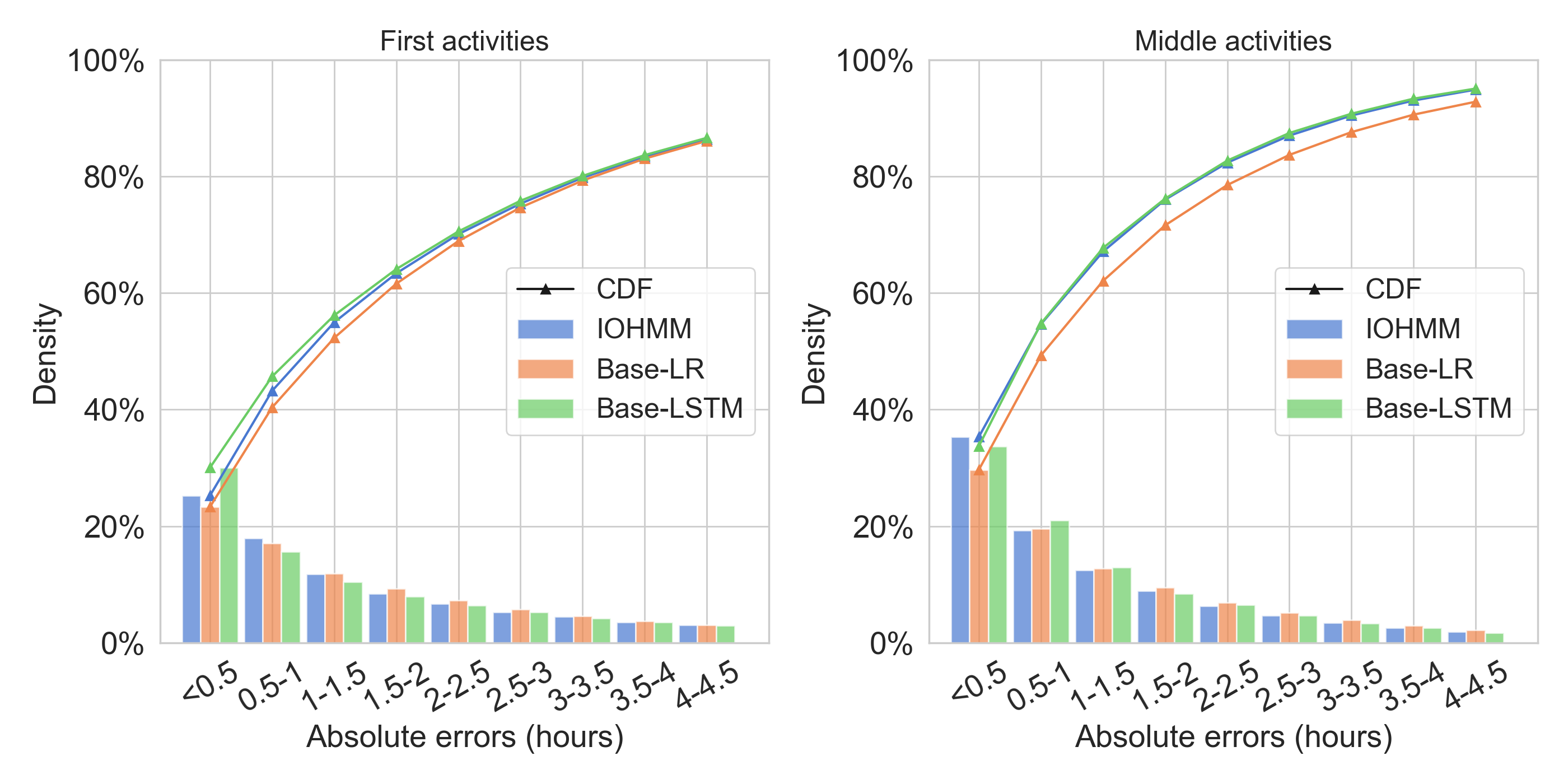}
\caption{Distribution of prediction errors for activity duration. The errors are aggregated in 0.5 hour intervals for better visualization. The solid lines are cumulative density functions (CDF) with colors corresponding to different models.}
\label{fig_duration_error}
\end{figure}

In addition to $R^2$, it is also useful to examine the magnitude of duration prediction errors. The distribution of absolute errors for duration prediction is shown in Figure \ref{fig_duration_error}. Overall, errors within 30 minutes account for the highest fraction for all models. For the middle activities, more than half of the activity duration can be predicted with errors within 1 hour for the IOHMM model. For the first activities, we observe that the LSTM model has a higher density in errors smaller than 30 minutes compared to IOHMM. However, for the middle activities, IOHMM accounts for a higher density for prediction errors within 30 minutes. This indicates that LSTM may have more advantages for the first activity duration prediction while IOHMM for middle activities. As for errors within 1.5 hours, the performance of IOHMM and LSTM models are similar, and both outperform the LR model. 

\begin{figure}[htb]
\centering
\includegraphics[width=1 \linewidth]{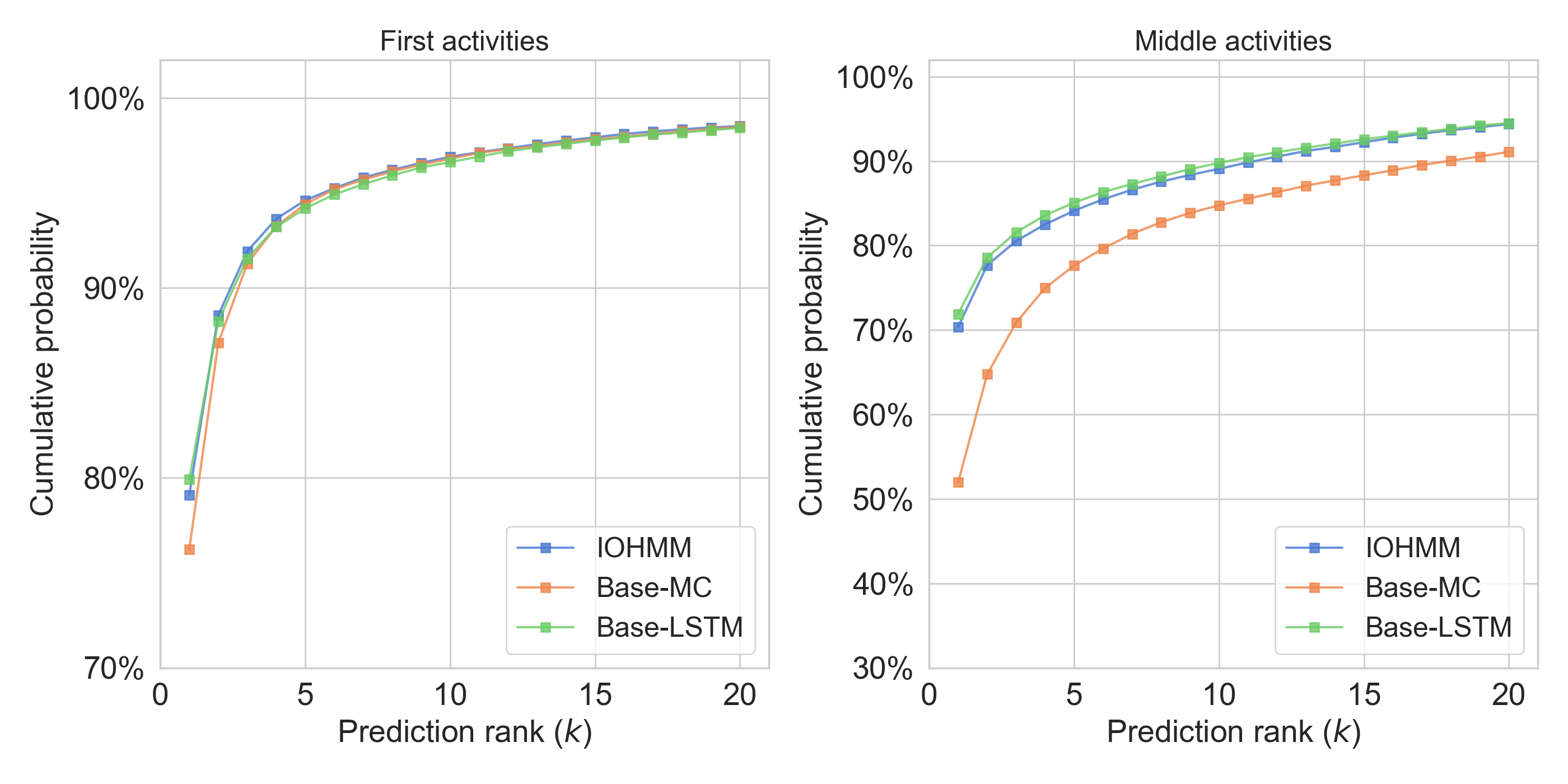}
\caption{Cumulative distribution of the prediction ranks for activity end location}
\label{fig_location_error}
\end{figure}

Figure \ref{fig_location_error} shows the cumulative distribution of prediction rank for activity end location. The cumulative probability (on the y-axis) at rank $k$ represents the probability that the true activity end location is among the top-$k$ (on the x-axis) most likely outcomes predicted by the model. We observe that, for the first activity, there is more than 90\% probability that one of the top 3 predictions in the IOHMM model is correct. But for middle activities, we need to include the top 10 predicted outcomes to achieve 90\% probability. The results imply that the origins of the first trips (i.e. first activity end locations) are easier to predict with limited variations than those of following trips. Similarly, IOHMM and LSTM models achieve comparable (essentially the same) performance, and both consistently outperform the MC model. 

\subsection{Factors impacting individual mobility predictability}
As shown in Figure \ref{fig_pred}, the prediction performance varies greatly among passengers. Hence, it is worth evaluating which attributes affect the individual's predictability. We estimated two linear regression models with the $R^2$ (for duration prediction) and prediction accuracy (for location prediction) of IOHMM as dependent variables. Independent variables are factors related to a user's travel frequency, regularity, fare card type, and inferred home location. Table \ref{tab_predictability} shows the results of estimated coefficients. We observe that the number of days with travel is significantly positive to location prediction, which implies that longer historical trips can increase the location predictability. Similarly, the mean number of trips per day is significantly positive for both duration and location prediction. This may be because a high mean number of trips per day reflects longer daily travel sequences, which can potentially make it easier to uncover sequential dependencies and ultimately help with prediction performance. Variables that indicate high travel irregularity, such as the standard deviation (std.) of travel frequency and departure time, have significant negative effects on the prediction performance. We also find that senior passengers' activity duration is harder to predict. In addition, the activity locations for users living in New Territories (one of the three main regions of Hong Kong, alongside Hong Kong Island and the Kowloon Peninsula) are easier to predict. This may be because New Territories is further away from the commercial business center of Hong Kong, and its residents generally have less diverse socioeconomic activities outside commuting between home and work. 

\begin{table}[H]
\caption{Factors on individual mobility predictability}
\centering
\begin{tabular}{@{}lrlrl@{}}
\toprule
\multirow{2}{*}{Variables}                                                                   & \multicolumn{4}{c}{Coefficients}                       \\ \cmidrule(l){2-5} 
                                                                                             & Duration             &    & End location       &    \\ \midrule 
Intercept                                                                                    & 0.6472               & ** & 0.8448               & ** \\
\# of days with travel                                                                       & 1.40$\times 10^{-5}$ &    & 8.83$\times 10^{-5}$ & **   \\
Mean \# of trips per day                                                                     & 0.1170               & ** & 0.0978               & ** \\
Std. \# of trips per day                                                                     & -0.1301              & **  & -0.2885              & **  \\
\begin{tabular}[c]{@{}l@{}} Std. of departure time of \\ the first trip in a day\end{tabular} & -0.0010              & ** & -0.0006              & ** \\
Student                                                                                      & -0.0345              &    & -0.0105               &    \\
Senior                                                                                       & -0.0739              & *  & 0.0068              &    \\
$^1$Living in Hong Kong Island                                                               & 0.0026              &   & 0.0088              &    \\
$^1$Living in New Territories                                                                 & 0.056              &   & 0.0184          &*    \\\bottomrule

\multicolumn{5}{l}{
\begin{tabular}[c]{@{}l@{}} Number of observations: 500. \\
Duration $R^2$: 0.287; End location $R^2$: 0.462 \\
$^{**}$: $p$-value $<0.01$; $^{*}$: $p$-value $<0.05$.\\
$^1$: A user's home location is inferred as the most frequently used tap-in \\station for the first trip of a day.
\end{tabular}} 
\end{tabular}
\label{tab_predictability}
\end{table}

\subsection{Latent activity identification}\label{sec_latent_act}
\begin{figure}[htb]
\centering
\includegraphics[width=1 \linewidth]{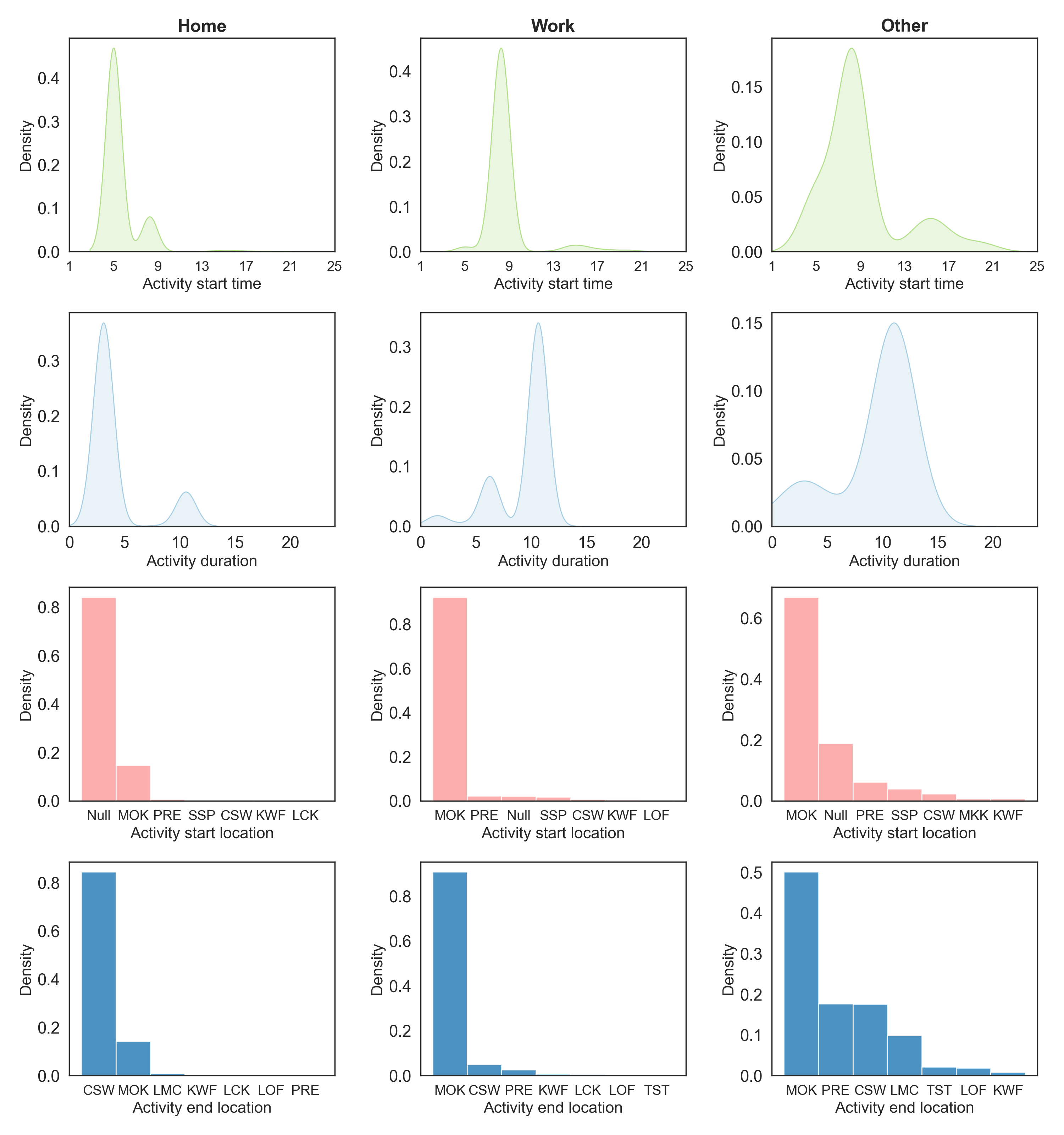}
\caption{Activity patterns of a selected individual with $N^u$ = 3. The $i$-th column represents the conditional distributions $\mathbb{P}(\cdot \;|\; A = i)$. Each column (i.e. each latent activity) is associated with a semantic label (i.e. ``home'', ``work'', and ``other'') based on its distribution patterns.}
\label{fig_individual_pattern}
\end{figure}

Although IOHMM has similar performance as the LSTM model, the advantage of IOHMM is its ability to identify latent activity patterns for each individual. Figure \ref{fig_individual_pattern} presents the spatiotemporal distributions of three latent activities (see Section \ref{sec_interp}) for a selected individual. Note that the activity labels are manually assigned based on their corresponding characteristics.

The first activity (i.e. the first column) has a start time peak around 4:00 AM, a duration peak around 4 hours, a dominant activity start location ``Null'' (i.e. the one with the highest probability and is much higher than others), and a dominant activity end location ``CSW'' (Cheung Sha Wan). This is obviously associated with ``home'' activity because people usually stay at home from the beginning of a day (4:00 AM) to the departure time for working (around 8:00 AM) with a duration of 4 hours. By definition, the first activity of a day has no activity start location (i.e. Null). And the dominant activity end location (i.e. CSW) should be the nearest station to his/her house. 

The second activity (i.e. the second column) has a start time peak around 8:00 AM, a duration peak around 10 hours, a dominant activity start and end location ``MOK'' (Mong Kok). We can easily associate it with the ``work'' activity, because the activity start time and duration match the typical work schedule, and the activity start and end locations are the same, which means during the activity the user does not move. MOK station is located in the CBD area in Hong Kong, which should be the nearest station to his/her working place. 

The third activity (i.e. the third column) has a relatively dispersed start time and duration compared to the first two. And the activity locations are more diverse. So, we associated it with ``other'' activities such as entertainment and dining. The activity start and end locations with the highest probability are both MOK. It seems the user prefers to perform other activities around MOK as well, which makes sense as MOK is one of the busiest areas in Hong Kong with many shops and restaurants. It is also likely that some of the other activities are work-based, as they may be planned around the work location.

\begin{figure}[htb]
\centering
\includegraphics[width=0.5 \linewidth]{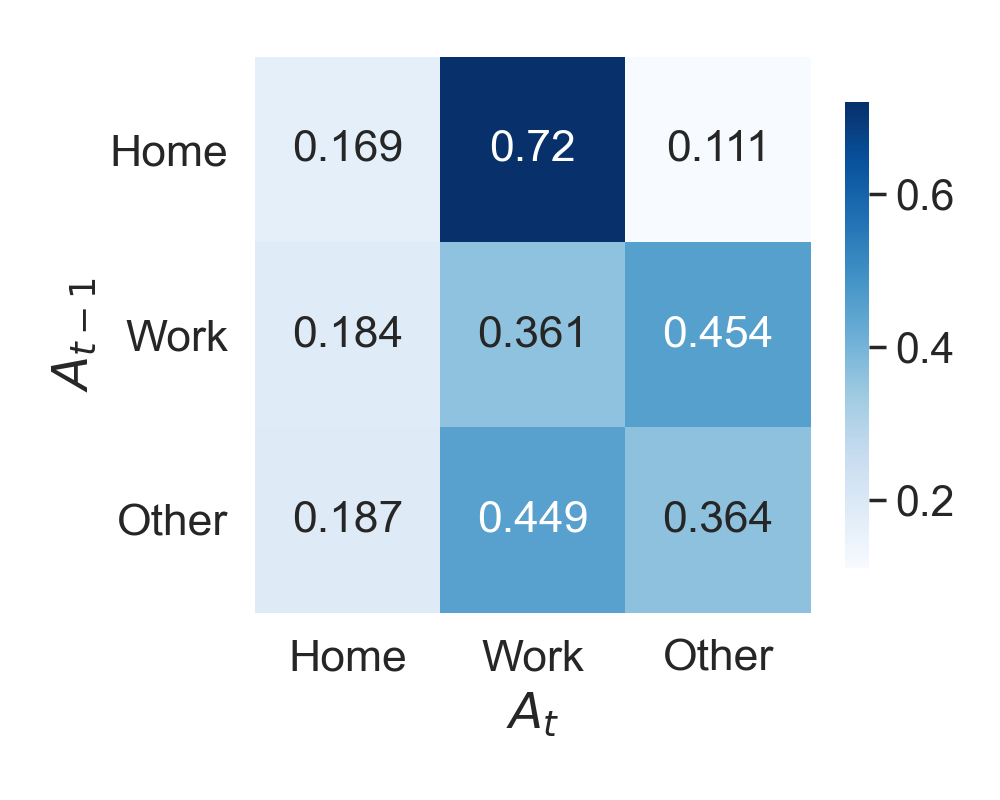}
\caption{Activity transition matrix ($\mathbb{P}(A_t\;|\;A_{t-1})$) of the selected individual}
\label{fig_individual_transition}
\end{figure}

The activity transition matrix is shown in Figure \ref{fig_individual_transition}. As expected, the transition from home to work shows the highest probability. It is worth noting that as the activity after the last trip in a day is omitted from analysis (see Section \ref{sec_act_based}), the typical work to home transition is not revealed in the model. The most likely activity following work is other, which may represent work-based shopping, dining, and entertainment activities (corresponding to results in Section \ref{sec_latent_act}). And the most likely activity following other is work. This also makes sense because the user usually conducts work-based other activities (such as dining) and after that he/she may need to return to work. 

Since the mean activity duration is specified with a linear model in IOHMM (see Section \ref{model_speciciation}), the estimated parameters in the linear model are useful for understanding other contextual factors affecting activity duration. Table \ref{tab_duration_para} summarizes some estimated parameters with interpretability for the same selected individual. As expected, the duration for all activities is higher on rainy days (compared to those without rain). It is worth noting that the sum of three activity duration is not a fixed value, since the activity after the last trip is not considered. Thus, the rainy parameter can be positive for all three activities, which represents users may delay their trip departure time when it rains outside. Monday has a positive impact on the work and other activities duration (the impact of Monday on home activity is negligible compared to the other two). Sunday has a negative impact on the work activity duration and a positive impact on home activity duration. On public holidays, the duration of home activities increases and decreases for the other two. Since ``other'' activities for this individual is usually work-based, all these effects are reasonable. 

\begin{table}[htb]
\caption{Estimated parameters for duration prediction}
\centering
\begin{tabular}{@{}ccccc@{}}
\toprule
\multirow{2}{*}{Activity} & \multicolumn{4}{c}{Estimated parameters ($\bm{\theta_{emr,i}}$)}  \\\cmidrule(l){2-5} 
                          & Rainy     & Monday    & Sunday    & Public holiday \\ \midrule
Home                      & 0.159  & 0.001  & 0.025  & 0.174        \\
Work                      & 0.336  & 0.173  & -0.221  & -0.065        \\
Others                     & 0.364  & 0.211  & -0.001   & -0.013        \\ \bottomrule
\end{tabular}
\label{tab_duration_para}
\end{table}

\section{Conclusion and discussion}
This paper proposes an IOHMM framework to simultaneously predict the time and location of an individual's next trip using smart card data. The prediction task can be transformed into predicting the hidden activity duration and end location, which enables a natural behavioral representation. Based on a case study with data from Hong Kong's MTR system, we show that the proposed IOHMM model has similar prediction performance as the advanced LSTM model, and significantly outperforms the benchmark models. Unlike LSTM, the proposed IOHMM model can also be used to analyze hidden activity patterns, which provides meaningful behavioral interpretation for why an individual makes a certain trip. Therefore, the activity-based prediction framework offers a way to combine the predictive power of machine learning methods and the behavioral interpretability of activity-based models. The estimated activity (or travel purpose) information can facilitate the development of situational awareness in intelligent transportation applications, such as personalized traveler information.

Future work can improve and extend the proposed approach. First, due to limitations of transit smart card data, only individual public transit trips are observed. This causes an activity defined in this study to potentially include trips made with alternative modes of transportation (such as taxi trips), introducing challenges in activity identification and interpretation. However, the proposed framework is not constrained to transit smart card data. Future research may fuse multiple data sources (such as smart cards, GPS, and cell phone data) to construct complete user trajectories. Second, an individual-based modeling framework may not work well when the observed trajectories are limited (e.g., infrequent or new users). Future studies may leverage user clustering techniques to extract similar users' travel patterns as additional input \cite{alhasoun2017city}, which can compensate for the sparsity of individual data.

\appendices
\section{Hyper-parameter space of the LSTM model}\label{sec_hyper}
The hyper-parameters of the LSTM model used in this study (for all individuals) are $M = 1$, $K = 50$, dropout rate $=0.3$, $l_1$ regularization $=0$, $l_2$ regularization $=0$, batch size $=30$. The model is trained using Adam optimizer (with the default learning rate) with 200 training epochs. 

\begin{table}[H]
\caption{Hyper-parameter space of the LSTM model}
\centering
\begin{tabular}{@{}ll@{}}
\toprule
Hyper-parameters                  & Value space                  \\ \midrule
\# of LSTM layers $M$         & \{1, 2, 3, 4, 5\}                \\
\# of units in LSTM layer $K$ & \{30, 50, 100, 150, 200\}        \\
Dropout rate                  & \{0.1, 0.3, 0.5, 0.7\}          \\
$l_1$ regularization          & \{0, $10^{-6}$, $10^{-4}$, 0.01, 0.1, 0.5\} \\
$l_2$ regularization          & \{0, $10^{-6}$, $10^{-4}$, 0.01, 0.1, 0.5\} \\
Batch size                    & \{20, 30, 50, 70\}              \\ \bottomrule
\end{tabular}
\label{tab_LSTM_space}
\end{table}

\section*{Acknowledgment}

The authors would like to thank Hong Kong's Mass Transit Railway (MTR) for their support and data availability for this research.

\ifCLASSOPTIONcaptionsoff
  \newpage
\fi


\bibliographystyle{IEEEtran}
\bibliography{mybib}

\begin{IEEEbiography}[{\includegraphics[width=1in,height=1.5in,clip,keepaspectratio]{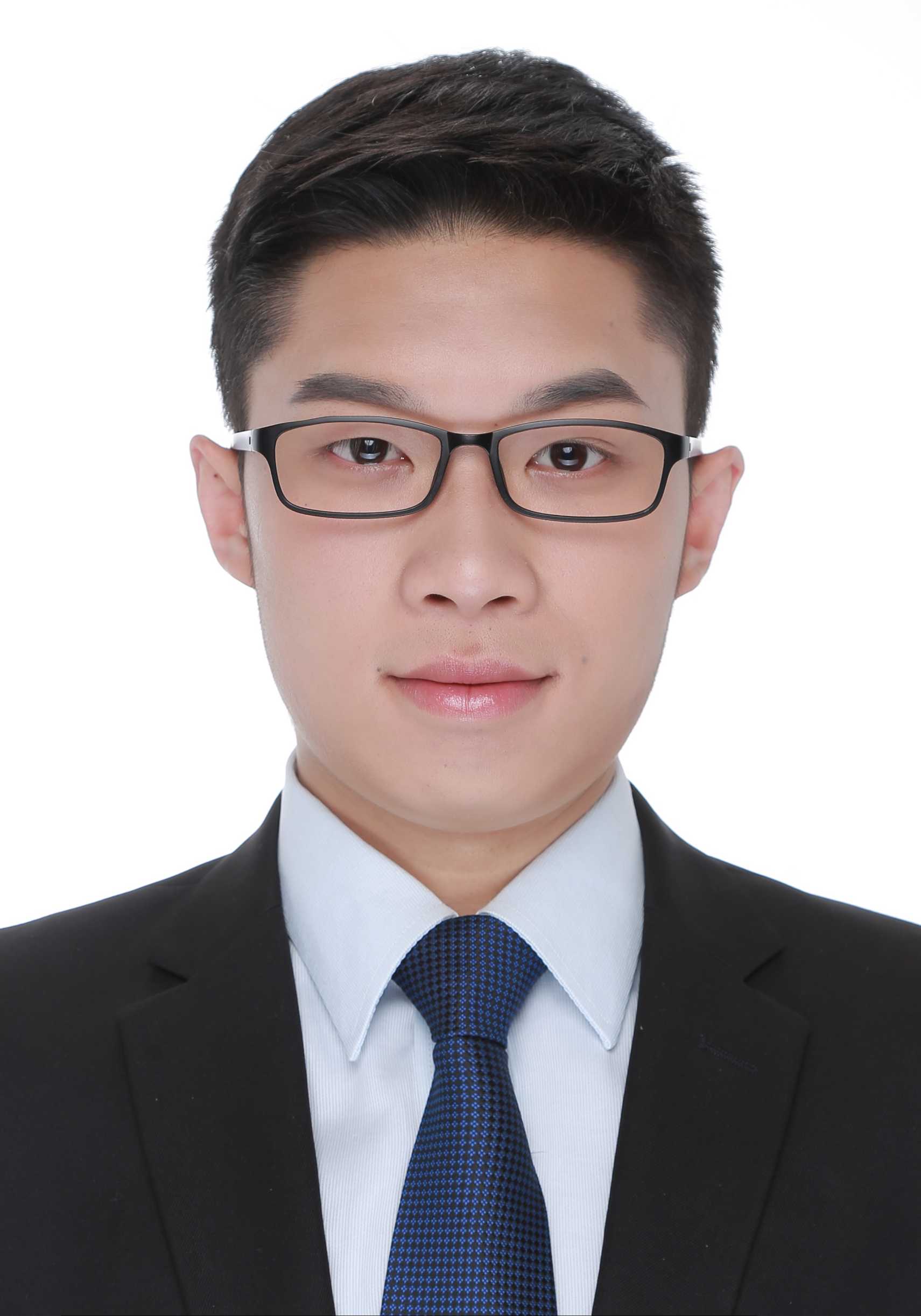}}]{Baichuan Mo} is a Ph.D. candidate in the Department of Civil and Environmental Engineering at MIT. He holds a dual Master's degree in Transportation and Computer Science from MIT, and a Bachelor's degree in Civil Engineering from Tsinghua University. His research interests include data-driven transportation modeling, demand modeling, and applied machine learning, with a specific application in public transit systems. 
\end{IEEEbiography}

\begin{IEEEbiography}[{\includegraphics[width=1in,height=1.5in,clip,keepaspectratio]{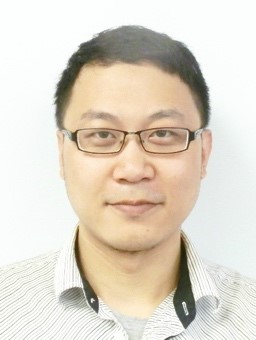}}]{Zhan Zhao} is an Assistant Professor in Department of Urban Planning and Design at the University of Hong Kong (HKU). He holds a Ph.D. degree from the Massachusetts Institute of Technology, a Master's degree from the University of British Columbia, and a Bachelor's degree from Tongji University. Prior to joining HKU, he was a senior data scientist at Via Transportation, Inc. His research interests include human mobility, public transportation systems, and urban data science. 
\end{IEEEbiography}

\begin{IEEEbiography}[{\includegraphics[width=1in,height=1.5in,trim = 5mm 15mm 5mm 5mm,clip,keepaspectratio]{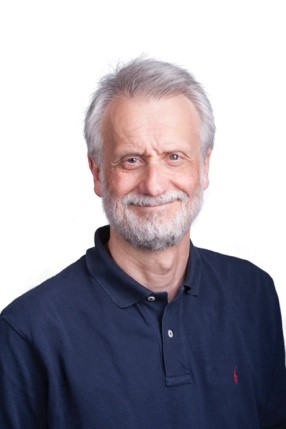}}]{Haris N. Koutsopoulos} is Professor in the Department of Civil and Environmental Engineering at Northeastern University in Boston and Guest Professor at KTH Royal Institute of Technology in Stockholm. His current research focuses on the use of data from opportunistic and dedicated sensors to improve planning, operations, monitoring, and control of urban transportation systems, including public transportation. He founded the iMobility lab, which uses Information and Communication Technologies to address urban mobility problems. The lab received the IBM Smarter Planet Award in 2012.
\end{IEEEbiography}

\begin{IEEEbiography}[{\includegraphics[width=1in,height=1.5in,trim = 11mm 35mm 11mm 5mm,clip,keepaspectratio]{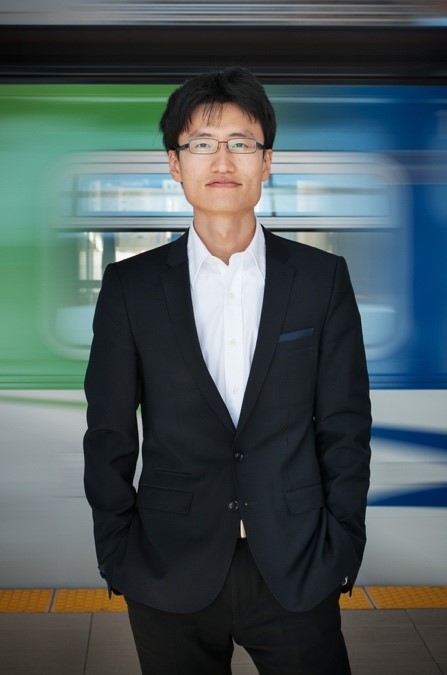}}]{Jinhua Zhao} is the Edward H. and Joyce Linde Associate Professor of City and Transportation Planning at MIT. He brings behavioral science and transportation technology together to shape travel behavior, design mobility systems, and reform urban policies. Prof. Zhao directs the MIT Urban Mobility Lab and Public Transit Lab. 
\end{IEEEbiography}

\end{document}